\pdfoutput=1

\documentclass[11pt]{article}

\usepackage{ACL2024}

\usepackage{times}
\usepackage{latexsym}

\usepackage[T1]{fontenc}

\usepackage[utf8]{inputenc}

\usepackage{microtype}

\usepackage{graphicx}
\usepackage{multicol}
\usepackage{multirow}
\usepackage{amsmath}
\usepackage{amssymb}
\usepackage{bbding}
\usepackage{subfigure}
\usepackage{colortbl}
\usepackage{pifont}
\usepackage{bbm}
\usepackage{makecell}
\usepackage{bbding}
\usepackage{mathrsfs}
\usepackage{bm}
\usepackage{CJKutf8}
\usepackage{booktabs}
\usepackage{tabu}
\usepackage{xcolor}
\usepackage{colortbl}

\definecolor{Gray}{gray}{0.93}

\newcommand{\ourmethod}{\textsc{UniCoder}}
\newcommand{\instruct}{\textsc{UniCoder-Instruct}}
\newcommand{\benchmark}{\textsc{UniCoder-Bench}}
\newcommand{\uot}{UoT}
\newcommand{\uc}{\textbf{\texttt{UniCode}}}
\NewDocumentCommand\emojiowl{}{
$\vcenter{\hbox{\includegraphics[height=1.5em]{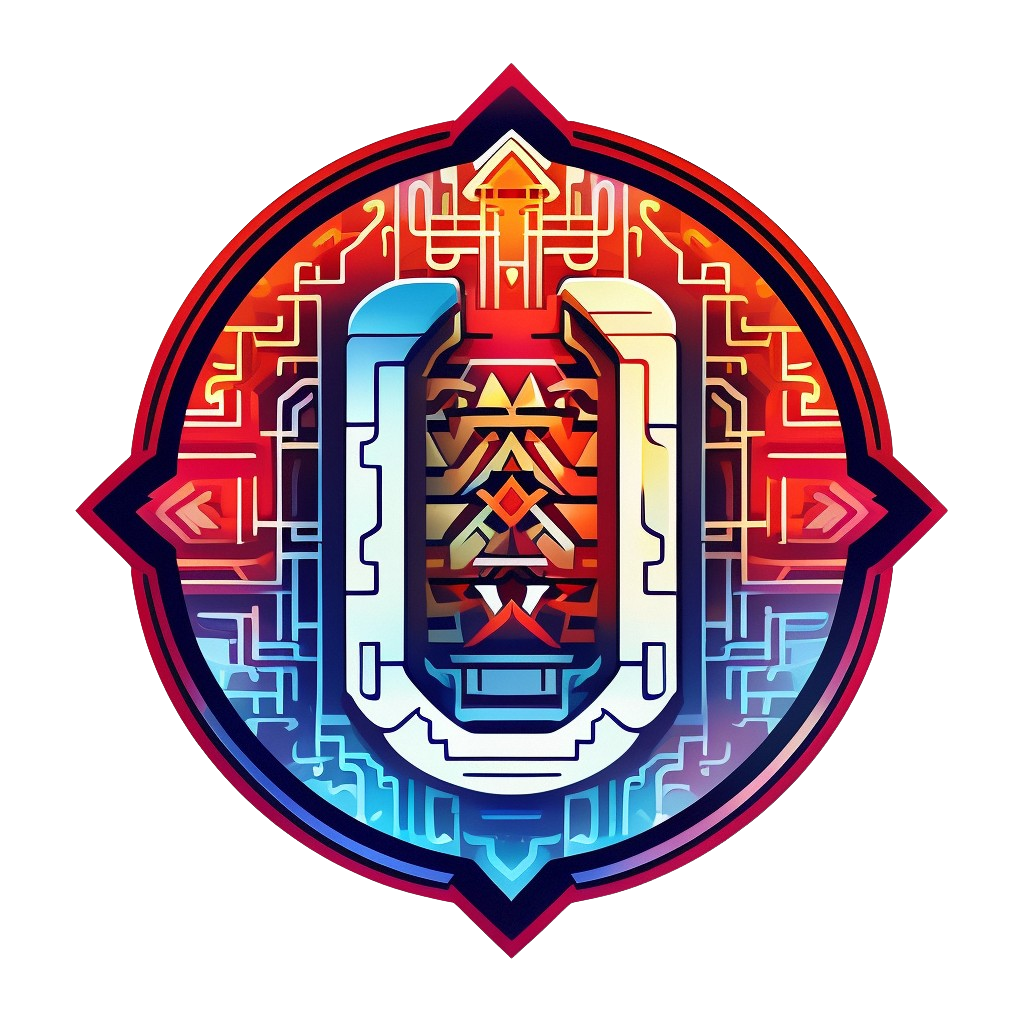}}}$
}

\newcommand{\greentick}{{\color{green}\ding{51}}}
\newcommand{\redcross}{{\color{red}\ding{55}}}

\title{\ourmethod{}\emojiowl{}: Scaling Code Large Language Model via Universal Code}

\author{
  Tao Sun\textsuperscript{\rm 1 \thanks{\ \ Equal contribution.}},
  Linzheng Chai\textsuperscript{\rm 1 *},
  Jian Yang\textsuperscript{\rm 1 *}\thanks{\ \ Corresponding Author.},
  Yuwei Yin\textsuperscript{\rm 2}, 
  Hongcheng Guo\textsuperscript{\rm 1}, \\
  {\bf Jiaheng Liu}\textsuperscript{\rm 1}, 
  {\bf Bing Wang}\textsuperscript{\rm 1},
  {\bf Liqun Yang}\textsuperscript{\rm 1}, 
  {\bf Zhoujun Li}\textsuperscript{\rm 1} \\
  \textsuperscript{\rm 1}State Key Laboratory of 
Complex \& Critical Software Environment, Beihang University; \\
  \textsuperscript{\rm 2}Department of Computer Science, University of British Columbia \\
  \{buaast, challenging, jiaya, hongchengguo\}@buaa.edu.cn; \\
  \{liujiaheng, bingwang, lqyang, lizj\}@buaa.edu.cn; yuweiyin@cs.ubc.ca \\
}

\begin{document}

\maketitle

\begin{abstract}
Intermediate reasoning or acting steps have successfully improved large language models (LLMs) for handling various downstream natural language processing (NLP) tasks.
When applying LLMs for code generation, recent works mainly focus on directing the models to articulate intermediate natural-language reasoning steps, as in chain-of-thought (CoT) prompting, and then output code with the natural language or other structured intermediate steps.
However, such output is not suitable for code translation or generation tasks since the standard CoT has different logical structures and forms of expression with the code.
In this work, we introduce the universal code (\uc{}) as the intermediate representation. It is a description of algorithm steps using a mix of conventions of programming languages, such as assignment operator, conditional operator, and loop.
Hence, we collect an instruction dataset \instruct{} to train our model \ourmethod{} on multi-task learning objectives. \instruct{} comprises natural-language questions, code solutions, and the corresponding universal code.
The alignment between the intermediate universal code representation and the final code solution significantly improves the quality of the generated code. The experimental results demonstrate that \ourmethod{} with the universal code significantly outperforms the previous prompting methods by a large margin, showcasing the effectiveness of the structural clues in pseudo-code.\footnote{\url{https://github.com/ASC8384/UniCoder}}
\end{abstract}

\section{Introduction}
\label{sec:introduction}

The field of code translation and generation has advanced significantly~\cite{code_translation_compiler,CodeTransOcean} with the advent of code-specific large language models (LLMs). Code LLMs, such as StarCoder~\cite{starcoder} and Code-Llama~\cite{code_llama}, are capable of generating executable code by analyzing natural language prompts.
Chain-of-thought (CoT) prompting~\cite{cot} has emerged as the leading technique in enhancing LLMs, where the intermediate steps provide a structured pathway from the problem statement to the solution, effectively mirroring the human problem-solving process.

\begin{figure}[t]
\centering
\includegraphics[width=1.0\linewidth]{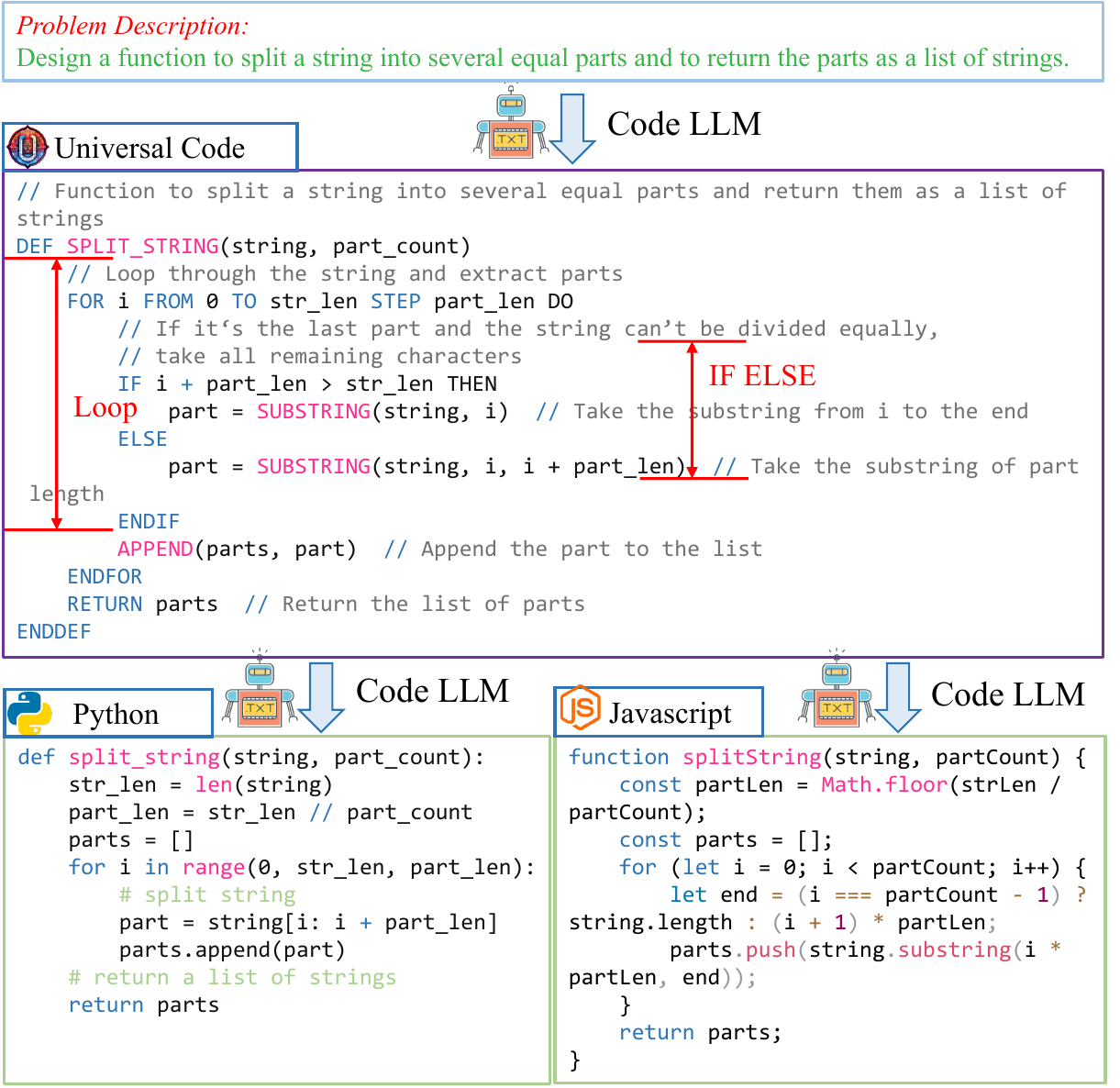}
\caption{An example of \ourmethod{}. The Code LLM solves the code generation question by ``translating'' the pseudocode description (Universal Code) into executable code of the target programming language.}
\vspace{-10pt}
\label{fig:intro}
\end{figure}

Considering the low accuracy of CoT in coder generation, structure CoT (SCoT)~\cite{scot} is proposed to minimize the gap between the intermediate steps and the generated code.
More intuitively, using a universal code as the intermediate representation to handle multiple programming languages (PL) is promising.
Here, universal code is a blueprint for implementing an algorithm, which helps to make the design of algorithms logically clear and readily comprehensible. Moreover, it is universal across different programming languages (PL-agnostic) since it typically does not follow specific syntax and omits execution details.
Yet, \textit{how the universal code is used for code translation and generation in multilingual scenarios remains underexplored.}

In this work, we scale up the code LLMs to support multiple programming languages via the universal code (\uc{}), which is used as an efficient and language-independent intermediate representation of the key algorithm principles.
Specifically, we first define \uc{} by specifying grammar rules and providing paradigms, followed by prompting GPT-4~\cite{gpt4} to create an instruction dataset \instruct{} comprising natural-language questions, code solutions, and the corresponding universal code, as shown in Figure~\ref{fig:intro}.
Then, the \ourmethod{} model is built by performing instruction tuning~\cite{flan} on multi-task learning objectives, including zero-shot question-answer generation (question$\to$code), question-universal-code generation (question$\to$\uc{}$\to$code), universal-code-solution translation (\uc{}$\to$code), and Universal-code-of-Thought (UoT) objectives.
In UoT, the model is required to generate the universal code before the executable code.

\ourmethod{} is evaluated on the Python benchmark (Humaneval~\cite{codex} and MBPP~\cite{mbpp}) and the extended multilingual benchmark MultiPL-E~\cite{multiple}. The results demonstrate that \ourmethod{} consistently achieves state-of-the-art performance across all languages, notably surpassing the previous baselines.
Furthermore, the ablation study verifies the efficacy of the proposed method, and extra discussions provide insights into the effect of our method.
The contributions are summarized as follows:
\begin{itemize}
\setlength\itemsep{0em}
    \item We introduce the universal code \uc{}, which is agnostic to programming languages, allowing LLMs to grasp the essence of algorithms step by step. In addition, the instruction dataset \instruct{} is collected and provided for follow-up research.
    \item We propose \ourmethod{}, a code generation method that uses multi-task learning objectives to fine-tune the code LLMs with the help of \uc{}. The objectives include question-answer generation (QA), question-universal-code generation (QP), universal-code-answer translation (PA), and Universal-code-of-Thought (UoT).
    \item As extensive experiments show, our method \ourmethod{} consistently outperforms the previous baselines on different benchmarks, including HumanEval, MBPP, and MultiPL-E. To further verify the effectiveness of the universal code, we propose \benchmark{} to test the capabilities of code LLMs.
\end{itemize}

\section{\instruct{}}
\label{sec:unicoder_instruct}

\begin{figure}[t!]
\centering
\includegraphics[width=0.85\columnwidth]{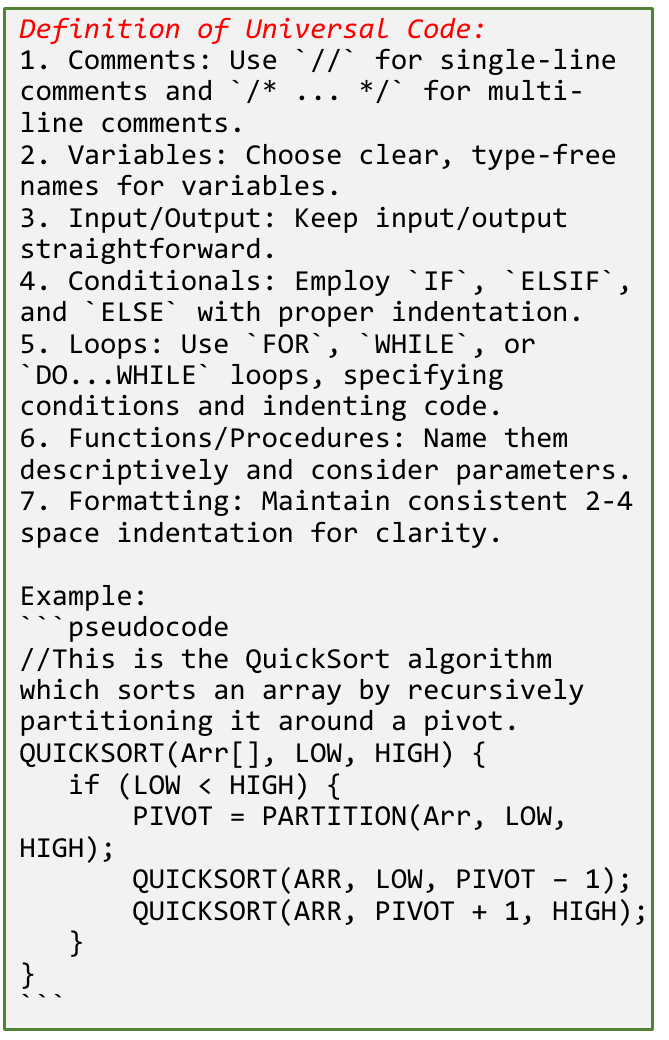}
\caption{Definition of the universal code.}
\vspace{-5pt}
\label{fig:universal_code}
\end{figure}

\begin{figure}[t!]
\centering
\includegraphics[width=0.85\columnwidth]{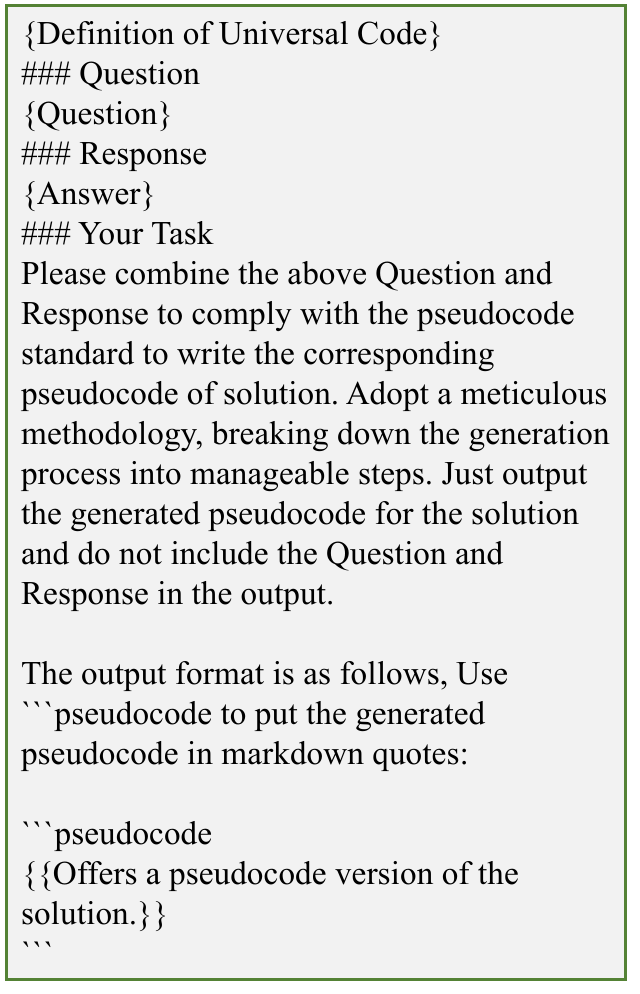}
\caption{Prompt of generating \uc{}.}
\vspace{-5pt}
\label{fig:prompt_qa2u}
\end{figure}

\paragraph{Definition of Universal Code.}
Universal code is designed for expressing algorithms in a form that is easily understood by humans, blending programming language syntax with natural language descriptions and mathematical notation to outline the steps of an algorithm without the complexity of full coding details. It omits machine-specific implementations to focus on the core logic, making it a popular choice for documentation in educational materials and the preliminary design phases of software development. By abstracting away from the intricacies of actual code, pseudocode facilitates clear communication of algorithmic concepts across various programming environments.
The definition of the universal code, as shown in Figure~\ref{fig:universal_code}, is based on the following principles:
\begin{itemize}
\setlength\itemsep{0em}
    \item \textbf{Comments}: Provide explanations and context for code segments, making it easier for others to understand the intent and functionality.
    \item \textbf{Variables}: Enhance code readability and maintainability by using meaningful names that convey the purpose of the variables without relying on data type specifications.
    \item \textbf{Input/Output}: Simplify the interaction with data entering and leaving the system, ensuring these operations are clear and easy to trace.
    \item \textbf{Conditionals}: Clarify decision-making processes within the code by using structured and indented conditional statements that define clear execution paths.
    \item \textbf{Loops}: Facilitate the repetition of code blocks in a controlled manner, with clearly defined start and end conditions, making the iterative processes understandable.
    \item \textbf{Functions/Procedures}: Increase modularity and reusability by naming functions and procedures descriptively, and by using parameters effectively to encapsulate functionality.
    \item \textbf{Formatting}: Improve the overall visual organization of the code by applying consistent indentation, which helps in delineating hierarchical structures and logical groupings within the code.
\end{itemize}

\paragraph{Construction From Instruction Dataset.}\label{para:gen_unicode}

For a programming language $L$, given the existing code instruction pair $(q_{\alpha},a_{\alpha}) \in D_{s}^{L}$, where $q_{\alpha}$ and $a_{\alpha}$ are question and answer from $D_{s}^{L}$, we create the universal code instruction dataset $D_{u_{\alpha}}^{L}$ by prompting LLMs to generate the universal code $p_\alpha$ and then add $(q_{\alpha},a_{\alpha},p_{\alpha})$ into $D_{u_{\alpha}}^{L}$.
Figure~\ref{fig:universal_code} shows the definition of the code universal and Figure~\ref{fig:prompt_qa2u} is the prompt for LLMs to generate \uc{}. \textcolor{orange}{\{Definition of Universal Code\}}, \textcolor{blue}{\{Question\}}, and \textcolor{teal}{\{Answer\}} denote the slots for definition of the universal code $p_{\alpha}$, the question of the instruction data $q_{\alpha}$, and the answer of the instruction $a_{\alpha}$, respectively.
Given $K$ different programming languages $L_{all} = \{L_{k}\}_{k=1}^{K}$, the multilingual programming instruction dataset with the universal code $D_{u_{\alpha}} = \{D_{u_{\alpha}}^{L_{k}}\}_{k=1}^{K}$ are created for supervised fine-tuning (SFT)~\cite{ouyang2022rlhf}. In this work, we adopt the open-source instruction dataset.

\paragraph{Construction From Code Snippets.}\label{para:gen_qa}
For the unsupervised data (code snippets) widely existing on many websites (e.g., GitHub), we also construct the instruction dataset with the universal code from raw code snippets.
Specifically, we ask the LLM to generate the question $q_{\beta}$ and the corresponding code answer $a_{\beta}$ pair based on the original code snippet $c$ using the prompt ``Please generate the self-contained question and answer based on the given code snippet''. Then, we generate \uc{} $p_{\beta}$ and construct $(q_{\beta},a_{\beta},p_{\beta})$ triplets the same way as in Paragraph~\ref{para:gen_unicode}. In addition, an LLM scorer is applied to filter out the low-quality $(q_{\beta},a_{\beta},p_{\beta})$ triplets.
Therefore, given raw code snippets of different programming languages $L_{k} \in \{L_{k}\}_{k=1}^{K}$, we can construct instruction dataset with the universal code $D_{u_{\beta}}=\{D_{u_{\beta}}^{L_{k}}\}_{k=1}^{K}$ directly from such unsupervised data.
Finally, we combine these two instruction datasets to obtain $D_{u}=D_{u_{\alpha}} \cup D_{u_{\beta}}$, where $D_{u}^{L_{k}}=D_{u_{\alpha}^{L_{k}}} \cup D_{u_{\beta}^{L_k}}$ for each program langauge $L_{k} \in L_{all}$.

\paragraph{Evaluation Task for Universal Code.}
To test the capability of the LLMs in generating \uc{} from questions and translating \uc{} into answers, we design a code reconstruction task for evaluation.
Given the code snippet $c$, we require the LLM to generate \uc{} $p$ and then translate it into the code $c'$. The evaluation metric is not the similarity between $c$ and $c'$ but whether the restored code $c'$ can pass the test cases.
We expand the HumanEval and MBPP datasets to create our benchmark \benchmark{} comprising $164$ HumanEval samples and $500$ MBPP test samples.

\begin{figure*}[t!]
\begin{center}
    \includegraphics[width=1.0\textwidth]{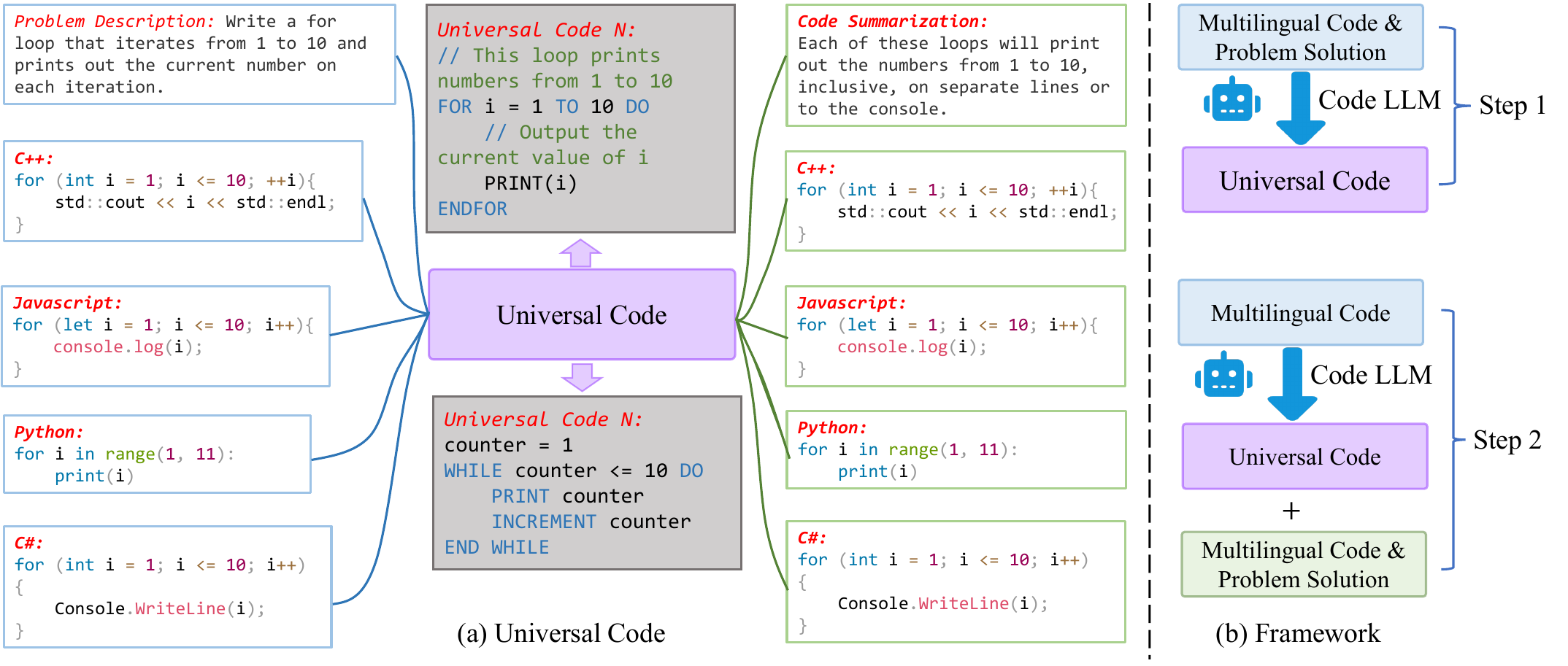}
    \caption{\textbf{Overview of \ourmethod{}}. (a) The function of universal code \uc{}; (b) The framework of our method \ourmethod{}. The universal code as the intermediate representation, our proposed framework can support code generation, code translation, and code summarization. In (a), the LLM encodes the code snippets of multilingual programming languages or the problem description questions into \uc {}. Then \uc{} is translated into the target output, i.e., the executable code of multilingual programming languages with a descriptive code summarization. In (b), we first ask the LLM to generate \uc{} with few-shot prompts. In the second stage, the instruction dataset, containing questions, answers, and \uc{}, is fed into the code LLM for fine-tuning.}
    \label{fig:model}
    \vspace{-10pt}
\end{center}
\end{figure*}

\section{\ourmethod{}}
\label{sec:unicoder}

\subsection{Model Overview}
In Figure~\ref{fig:model}, we first define the concept of the universal code with the essential components and then prompt the LLM to generate \uc{} $p$ based on the existing instruction data (questions $q$ and answers $a$) and the raw code snippets $c$. \uc{} is regarded as the intermediate representation for different tasks, including code generation, code translation, and code summarization. Our proposed model \ourmethod{} is trained on the instruction dataset $D_{u}$ with the multilingual objectives to fully unleash the potential of \uc{}.

\subsection{Code LLM with Universal Code}
Given the instructions dataset with $K$ multilingual programming languages $D_{u}=\{D_{u}^{L_{k}}\}_{k=1}^{K}$, the pre-trained code LLM $\mathcal{M}$ trained on $D_{u}$ can support Universal-code-of-Thought (\uot{}). It can be described as:
\begin{align}
    P(p,a|q) = P(p|q;\mathcal{M})P(a|q,p;\mathcal{M})
    \label{equ:p2q2a}
\end{align}
where $q$ (question) and $a$ (answer) are the instruction pair from $D_{u}$. Given the question $q$, the code LLM $\mathcal{M}$ first generates \uc{} $p$ and then outputs the final answer $a$, where $p$ provides key algorithm ideas with natural language comments.

\subsection{Multi-task Supervised Fine-tuning}
To fully unleash the potential of the \uc{}, we design multiple objectives to enhance the understanding and generation capability of code LLM.
\paragraph{Multi-task Fine-tuning.}
\begin{align}
    \mathcal{L}_{all} = \mathcal{L}_{qa} + \mathcal{L}_{qp} + \mathcal{L}_{pa} + \mathcal{L}_{uot}
    \label{equ:loss_all}
\end{align}where $\mathcal{L}_{qa}$ is the question-answer generation objective, $\mathcal{L}_{qp}$ is the question-universal-code generation objective, $\mathcal{L}_{pa}$ is the universal-code-answer translation objective, and $\mathcal{L}_{uot}$ is the Universal-code-of-Thought (\uot{}) objective.

Here, we introduce all four training objectives.
For all the following objectives, the multilingual corpora $D_{u}=\{D_{u}^{L_{k}}\}_{k=1}^{K}$ are given. $\mathcal{M}$ is the code LLM and $K$ is the number of programming languages.

\paragraph{Question-Answer Objective.}
The training objective $\mathcal{L}_{qa}$ of the standard instruction fine-tuning can be described as:
\begin{align}
    \mathcal{L}_{qa} = -\sum_{k=1}^{K} \mathbb{E}_{q,a \sim D_{u}^{L_{k}}} \left[ \log P(a|q; \mathcal{M}) \right]
    \label{equ:loss_qa}
\end{align}where $q$ and $a$ are the question and answer pair.

\paragraph{Question-Universal-Code Objective.}
The training objective $\mathcal{L}_{qp}$ of the auxiliary universal code generation task can be described as:
\begin{align}
    \mathcal{L}_{qp} = -\sum_{k=1}^{K} \mathbb{E}_{q,p \sim D_{L_{k}}} \left[ \log P(p|q; \mathcal{M}) \right]
    \label{equ:loss_qp}
\end{align}where $q$ and $p$ are the question and \uc{}.

\paragraph{Universal-Code-Answer Objective.}
The training objective $\mathcal{L}_{pa}$ of generating the executable code answer from \uc{} can be described as:
\begin{align}
    \mathcal{L}_{pa} = -\sum_{k=1}^{K} \mathbb{E}_{p,a \sim D_{L_{k}}} \left[ \log P(a|p; \mathcal{M}) \right]
    \label{equ:loss_pa}
\end{align}where $p$ and $a$ are \uc{} and the answer.

\paragraph{Universal-Code-of-Thought Objective.}
The training objective $\mathcal{L}_{uot}$ of generating \uc{} and then the executable code answer can be described as:
\begin{align}
    \mathcal{L}_{uot} = -\sum_{k=1}^{K} \mathbb{E}_{q,p,a \sim D_{L_{k}}} \left[ \log P(p,a|q; \mathcal{M}) \right]
    \label{equ:loss_uot}
\end{align}where $q$, $a$, and $p$ are the question, answer, and \uc{}, respectively.

\section{Experimental Setup}
\label{sec:experimental_setup}

\subsection{Instruction Dataset}
GPT-4 (\texttt{gpt-4-1106-preview})~\cite{gpt4} is used as the foundation model to generate the \instruct{}. We randomly extract code snippets within $1024$ tokens from the StarCoder dataset~\cite{starcoder} and let GPT-4 summarize the code snippets as the universal code. Based on each code snippet and the corresponding universal code, a self-contained coding problem with a correct solution is created.

\subsection{Baselines}

\paragraph{Proprietary Models.}
Based on a neural architecture known as generative pre-trained Transformers (GPT)~\cite{transformer,gpt}, GPT-3.5 and GPT-4 are LLMs trained on massive datasets of text, code, math equations, and more. They are also trained to follow instructions~\cite{ouyang2022rlhf}, which allows them to generate human-like responses. We use GPT-3.5 Turbo and GPT-4 as the proprietary models because they perform excellently in various code understanding and generation tasks.

\paragraph{Open-Source Models.}
To narrow the gap between open-source and closed-source models, a series of open-source models and instruction datasets are proposed to improve code LLMs and bootstrap their instruction-following ability. Starcoder~\cite{starcoder}, Code Llama~\cite{code_llama}, and DeepSeek-Coder~\cite{deepseek_coder} with different model sizes are introduced into the based model. OctoCoder~\cite{octopack}, WiazrdCoder~\cite{wizardcoder}, MagiCoder~\cite{magicoder}, and WaveCoder~\cite{wavecoder} are further fine-tuned on these based code LLMs.

\paragraph{Decontainmation.}
We apply data decontamination before training our \ourmethod{} models to decontaminate the code snippets from the starcoder data~\cite{starcoder}, by removing exact matches from HumanEval~\cite{codex}, MBPP~\cite{mbpp}, DS-1000~\cite{lai2023ds_1000}, and GSM8K~\cite{gsm8k}.

\subsection{Evaluation Benchmark}

\paragraph{HumanEval.} 
The HumanEval test set~\cite{codex} is a crafted collection of 164 Python programming problems to test the abilities of code generation models. For each problem, there are roughly 9.6 test cases to check whether the generated code works as intended. Humaneval has become one of the most popular benchmarks to measure how well these code-writing AI models perform, making it a key tool in the field of AI and machine learning for coding.

\paragraph{MBPP.}
The MBPP dataset~\cite{mbpp}, comprising approximately 1,000 Python programming challenges sourced from a crowd of contributors, is tailored for beginners in programming, focusing on core principles and the usage of the standard library. The MBPP test set comprised of $500$ problems is selected to evaluate the few-shot inference of the code LLMs. 

\paragraph{MultiPL-E.}
The MuliPL-E test set \cite{multiple} translates the original HumanEval test set to other 18 programming languages, i.e., Javascript, Java, Typescript, C++, and Rust. We use the MultiPL-E to evaluate the multilingual capabilities of the code LLMs.

\subsection{Evaluation Metrics}

\paragraph{Pass@k.} We adopt the Pass@k metric ~\cite{codex} to improve the reliability of our evaluation. We then count the total number of successfully passing test cases, denoted as $k$, to compute the Pass@k, thereby enhancing the accuracy and consistency of the performance assessment.
\begin{equation}
    \text{Pass@k} = \mathbb{E}\left[1 - \frac{ \binom{n}{k-c} }{ \binom{n}{k} } \right]
    \label{equ:pass_at_k}
\end{equation}where $n$ is the total number of generated samples for each problem, and $c$ is the number of correct generated code snippets passing all the test cases ($n > k \geq c$).

\subsection{Impletmentation Details}
We expand the open-source Evol-Instruct dataset \texttt{evol-code-alpaca-v1}~\cite{wizardlm} with nearly 110K samples into the instruction dataset with the universal code. For the code snippets collected from starcoderdata~\footnote{\url{https://huggingface.co/datasets/bigcode/starcoderdata}}, we choose 5K code snippets of each language (Python, Javascript, C++, Java, Rust, and Go) to construct the synthetic instruction dataset with universal code. Finally, we obtain the instruction dataset \instruct{} contains nearly 140K training samples.
Code-Llama and DeepSeek-Coder-Base are used as the foundational code LLMs for supervised fine-tuning (SFT). We fine-tune these foundation LLMs on nearly 150K samples generated from \texttt{evol-codealpaca-v1} and the starcoder pre-training data. \ourmethod{} is fine-tuned on \texttt{Standford\_Alpaca}\footnote{\url{https://github.com/tatsu-lab/stanford_alpaca}} with $8$ NVIDIA A100-80GB GPUs. The learning rate first increases into $8 \times 10^{-5}$ with $50$ warmup steps and then adopts a cosine decay scheduler. We adopt the Adam optimizer~\cite{adam} with a global batch size of $128$ samples, truncating sentences to $1536$ tokens.

\section{Results and Discussion}
\label{sec:results_and_discussion}

\begin{table*}[t!]
\centering
\resizebox{0.85\textwidth}{!}{
\begin{tabu}{lcccccc}
\toprule
\textbf{Models} & \textbf{Base Model} & \textbf{Params} & \textbf{Instruction Data} & \textbf{Model Weight} & \textbf{HumanEval} & \textbf{MBPP} \\ 
\midrule
\multicolumn{7}{l}{\textit{Proprietary Models}} \\
\arrayrulecolor{lightgray}\tabucline[0.4pt on 3pt off 3pt]{-} \addlinespace[0.1cm] 
GPT-3.5 & -  & -  &  -  &  -  & 72.6 & 81.6   \\
GPT-4   & -  & -  &  -  &  -  & 85.4 & 83.0   \\ \arrayrulecolor{black}\midrule
\multicolumn{7}{l}{\textit{Open-source Models}} \\
\arrayrulecolor{lightgray}\tabucline[0.4pt on 3pt off 3pt]{-} \addlinespace[0.1cm] 
StarCoder~\cite{starcoder}      &  -  & 15B   & \redcross{}   & \greentick{}   & 33.6     & 43.3       \\
WizardCoder~\cite{wizardcoder}  & StarCoder   &  15B  & \greentick{} &  \greentick{}  &   57.3   & 51.8  \\
OctoCoder~\cite{octopack}     &   StarCoder &  15B  &  \greentick{}  & \greentick{}   &   46.2   &  43.5  \\
WaveCoder-SC~\cite{octopack}     &   StarCoder &  15B  &  \greentick{}  & \greentick{}   &  50.5   &  51.0\\
\arrayrulecolor{lightgray}\tabucline[0.4pt on 3pt off 3pt]{-} \addlinespace[0.1cm] 
Code-Llama~\cite{code_llama}   &  -  &  7B  &  \redcross{}  & \greentick{}   & 33.5  &  41.4   \\
Code-Llama-Instruct~\cite{code_llama} & Code Llama  &  7B  &\greentick{}  &\greentick{} & 34.8 & 44.4\\
WaveCoder-CL~\cite{wavecoder}     &  Code Llama  &  7B  &  \greentick{}   & \greentick{}   &  48.1 & 47.2  \\
Magicoder-CL~\cite{magicoder}     &  Code Llama   &   7B & \greentick{}   & \greentick{}    & 60.4   &64.2 \\
\ourmethod{} (our method) &  Code Llama  &  7B  &  \greentick{} & \greentick{}   &  65.4    &  65.2      \\           \arrayrulecolor{lightgray}\tabucline[0.4pt on 3pt off 3pt]{-} \addlinespace[0.1cm] 
DeepseekCoder~\cite{deepseek_coder}  &  -  & 6.7B  &  \redcross{} & \greentick{}  & 49.4 & 60.6\\
WaveCoder-DS~\cite{wavecoder}     &Deepseek-Coder & 6.7B &  \greentick{} & \greentick{}& 64.0 &  62.8 \\
\textbf{\ourmethod{} (our method)} & Deepseek-Coder &  6.7B & \greentick{} &  \greentick{}   &  \textbf{70.6}          & \textbf{64.3}    \\
\arrayrulecolor{black}\bottomrule
\end{tabu}}
\caption{Evaluation results of Pass@$1$ on the HumanEval and MBPP benchmark. We use self-reported scores whenever available. All methods use greedy decoding and We use the reported scores of the previous work.}
\label{tab:coding_humaneval_mbpp}
\end{table*}

\begin{table*}[ht]
\centering
\resizebox{0.85\textwidth}{!}{
\begin{tabu}{@{}lccccccc|c@{}}
\toprule
\multirow{2}{*}[-\aboverulesep]{\textbf{Model}} & \multirow{2}{*}[-\aboverulesep]{\textbf{Params}} & \multicolumn{6}{c}{\textbf{Programming Language}}  & \\ \cmidrule(l){3-9}
             &   & Java & Javascript      & C++    & PHP      & Swift  & Rust & \textbf{Avg.}  \\ 
\midrule
\multicolumn{9}{l}{\textit{Proprietary models}}\\
\midrule
GPT-3.5  & -  & 69.2 & 67.1& 63.4 & 60.9& -& -   & - \\
GPT-4    & -  & 81.6 & 78.0& 76.4 & 77.2& -& -   & - \\
\midrule
\multicolumn{9}{l}{\textit{Open-source models}}\\
\arrayrulecolor{lightgray}\tabucline[0.4pt on 3pt off 3pt]{-} \addlinespace[0.1cm] \arrayrulecolor{black}
CodeLlama~\cite{code_llama}   & 34B  &  40.2 & 41.7 & 41.4 & 40.4 & 35.3 & 38.7  & 39.6  \\
CodeLlama-Python~\cite{code_llama} & 34B  &  39.5 & 44.7 & 39.1 &39.8 & 34.3 &39.7 & 39.5  \\
CodeLlama-Instruct~\cite{code_llama}        & 34B  &  41.5 & 45.9 & 41.5 & 37.0 & 37.6 &39.3 & 40.5  \\
WizardCoder-CL~\cite{wizardcoder}         & 34B   & 44.9 & 55.3 & 47.2 & 47.2 & 44.3 & 46.2 & 47.5  \\\arrayrulecolor{lightgray}\tabucline[0.4pt on 3pt off 3pt]{-} \addlinespace[0.1cm]\arrayrulecolor{black}
StarCoderBase~\cite{starcoder}         & 15B   &  28.5 & 31.7 & 30.6 & 26.8 & 16.7 & 24.5 &26.5  \\
StarCoder~\cite{starcoder}         & 15B   &  30.2 & 30.8 & 31.6 & 26.1 & 22.7 & 21.8 &27.2  \\
WizardCoder-SC~\cite{wizardcoder}  & 15B  &  35.8 & 41.9 & 39.0 & 39.3 & 33.7 & 27.1 &36.1  \\ \arrayrulecolor{lightgray}\tabucline[0.4pt on 3pt off 3pt]{-} \addlinespace[0.1cm]\arrayrulecolor{black}
CodeLlama~\cite{code_llama}   & 7B  &  29.3 & 31.7 & 27.0 & 25.1 & 25.6 & 25.5  & 27.4  \\
CodeLlama-Python~\cite{code_llama} & 7B  & 29.1 & 35.7 & 30.2 & 29.0 & 27.1 & 27.0 & 29.7  \\
\textbf{\ourmethod{} (Our method) }   & 7B   & \textbf{46.4} &\textbf{ 50.2} &\textbf{ 39.2}  & \textbf{40.4} &\textbf{41.2 }&\textbf{32.4} &\textbf{41.6} \\
\arrayrulecolor{black}\bottomrule
\end{tabu}}
\caption{Evaluation results of Pass@1 (\%) performance on the MultiPL-E benchmark. The baseline results are partly from the previous work~\cite{magicoder}.}
\label{tab:coding_multiple}
\end{table*}

\subsection{Main Results}
\label{subsec:results}

\paragraph{Python Code Generation.}
Table \ref{tab:coding_humaneval_mbpp} shows that \ourmethod{} significantly beats previous strong open-source baselines using \uot{}, closing the gap with GPT-3.5 and GPT-4. Magicoder~\cite{magicoder} and Wavecoder~\cite{wavecoder} both prove the effectiveness of instruction datasets from code snippets. Further, \ourmethod{} outperforms the WizardCoder with 15B parameters and Evol-Instruct techniques with the help of the \textbf{\uc{}}. 

\paragraph{Multilingual Code Generation.}
Table \ref{tab:coding_multiple} shows that \ourmethod{} significantly outperforms strong baselines CodeLlama and Starcoder. For the different backbones (Code Llama and Deepseek-Coder), our method beats most previous methods, especially in other languages, which demonstrates that \instruct{} can bring the capability of multilingual understanding and generation.

\subsection{Discussion}
\label{subsec:discussion}

\paragraph{Ablation Study.}
To verify the efficacy of each component, we conduct the ablation study step by step on HumanEval and MBPP. In Table~\ref{tab:ablation_humaneval_mbpp}, we observe that removing the multi-tasks objective (only keeping the \uot{} objective: Equation~\ref{equ:loss_uot}) will have a $-1.6$ performance drop in HumanEval and a $-1.3$ drop in MBPP. Removing \textbf{\uc{}} will further degrade the performance. The results support the effectiveness of each component of \ourmethod{}.

\begin{table}[t]
\centering
\resizebox{1.0\columnwidth}{!}{
\begin{tabular}{c|l|cc}
\toprule
\textbf{ID} & \textbf{Methods} &  \textbf{HumanEval}  & \textbf{MBPP} \\
\midrule
{\large{\ding{172}}} & \ourmethod{}    & 70.6 &  64.3 \\
{\large{\ding{173}}} & {\large{\ding{172}}} - Multi-tasks Objective & 67.4 & 60.2 \\
{\large{\ding{174}}} & {\large{\ding{173}}} - Universal Code        & 66.8 & 59.8 \\
\bottomrule
\end{tabular}
}
\caption{Ablation study of our proposed method on HumanEval and MBPP. \ourmethod{} is fine-tuned on the \instruct{} with the multi-task objectives.}
\label{tab:ablation_humaneval_mbpp}
\vspace{-10pt}
\end{table}

\paragraph{Effect on Universal Code.}
To discuss the effect of the different formats of the universal code, we use different definitions of universal code for \ourmethod{}. Specifically, we randomly sample 5K samples to generate the instruction dataset with different formats of \uc{}. 
\begin{itemize}
\setlength\itemsep{0em}
    \item \textbf{\uc{} 1}: It describes the naming conventions, variable declaration, operators, conditional statements, loops, and function structure that pseudocode should have.
    \item \textbf{\uc{} 2}: It separates the first set of standards and provides code examples for each, instead of applying them all together in the examples.
    \item \textbf{\uc{} 3}: It describes the code structure, variable rules, control structures, functions, comments, and assignment rules that pseudocode should have.
    \item \textbf{\uc{} 4}: It is similar to the first standard but specifies type-free names for variables.
    \item \textbf{\uc{} 5}: It provides an abstract, high-level architectural description, without setting standards for the code itself.
    \item \textbf{\uc{} 6}: It uses latex algorithm and algorithmic packages for description.
\end{itemize}

\begin{table}[t]
\centering
\resizebox{0.85\columnwidth}{!}{
\begin{tabular}{c|l|cc}
\toprule
\textbf{ID} & \textbf{Methods} &  \textbf{HumanEval}  & \textbf{MBPP} \\
\midrule
{\large{\ding{172}}} & \uc{} 1 & 53.2 & 51.5 \\
{\large{\ding{173}}} & \uc{} 2 & 52.8 & 51.2 \\
{\large{\ding{174}}} & \uc{} 3 & 53.5 & 50.5 \\
{\large{\ding{175}}} & \uc{} 4 & 53.8 & 49.5 \\
{\large{\ding{176}}} & \uc{} 5 & 49.5 & 50.2 \\
{\large{\ding{177}}} & \uc{} 6 & 48.2 & 48.4 \\
{\large{\ding{178}}} & \uc{} 1$\sim$4 & \bf 55.5 & \bf 52.2 \\
\bottomrule
\end{tabular}
}
\caption{Evaluation results of our method with different formats of the universal code.}
\label{effect_of_universal_code}
\vspace{-10pt}
\end{table}
In Table~\ref{effect_of_universal_code}, we can observe that the evaluation results of \uc{} 1$\sim$\uc{} 4 have better performance. Compared to the universal code format \uc{} 5 and \uc{} 6, \uc{} 1$\sim$\uc{} 4 has a clear definition and common structure, which brings more support for code generation.
Notably, the experiment {\large{\ding{178}}} performs the best by combing the training data of {\large{\ding{172}}}$\sim${\large{\ding{175}}}.
The experimental results show that the concrete definition of \uc{} and the combination of it can effectively improve the model performance.

\subsection{Code-\uc{}-Code}
To compare the capabilities of different code LLMs, we create a test set (denoted as \benchmark{}) by prompting the code LLM to generate \uc{} and translate it into the executable code.
We check the correctness of each translated code with the test cases, denoted as Pass@1 of the universal code.
Code-Llama-7B is fine-tuned on the Code Alpaca dataset and our dataset \instruct{} separately. The results of fine-tuned Code-Llama models on \benchmark{} are shown in Table~\ref{tab:unicoder_benchmark}. Our method \ourmethod{} is more accurate in passing the test cases than the Code-Llama baselines, demonstrating its excellent code understanding and generation abilities.
\begin{table}[t]
\centering
\resizebox{1.0\columnwidth}{!}{
\begin{tabular}{c|c|cc}
\toprule
\textbf{Method} & \textbf{Params} &  \textbf{Python}  & \textbf{Other Languages} \\
\midrule
Code-Llama-Instruct & 7B & 33.3  & 26.2    \\
Code-Llama-Alpaca   & 7B & 44.2  & 29.1    \\
\ourmethod{}        & 7B & 45.2  & 31.3    \\
\bottomrule
\end{tabular}
}
\caption{Pass@1 scores of our method \ourmethod{} and two Code-Llama baselines for Code-\uc{}-Code.}
\label{tab:unicoder_benchmark}
\vspace{-10pt}
\end{table}

\section{Related Work}
\label{sec:related_work}
\paragraph{Code Understanding and Generation.}
Code understanding and generation as the key tasks to substantially facilitate the project development process, including code generation~\cite{codex,mbpp,repocoder,chai2024mceval,deng2024r2c2}, code translation~\cite{code_translation_compiler}, automated testing~\cite{LLM_Testing}, bug fixing~\cite{octopack}, code refinement~\cite{code_refinement}, code question answering~\cite{code_qa}, and code summarization~\cite{Summarization}. Researchers \citet{ERNIE_Code} have undertaken extensive endeavors to bridge natural language and programming languages.
With less ambiguous prompt styles, \citet{Pseudo_Instructions} using pseudocode improves the performance of NLP tasks. \citet{Pseudo_2015} uses traditional machine learning to achieve code to pseudocode conversion. \citet{PrototypingMaker} also shows that designers and programmers can speed up the prototyping process, and ground communication between collaborators via prompt-based prototyping. To verify that the generated code is correct, there are some code synthesis evaluation frameworks, including EvalPlus~\cite{evalplus}, HumanEval~\cite{codex}, HumanEval-X~\cite{codegeex}, and MBPP~\cite{mbpp}. 

\paragraph{Large Language Models for Code.}
Since CodeBERT~\cite{code_bert} first connected code tasks with pre-trained models, large language models for code have developed rapidly, demonstrating extraordinary performance on almost all code tasks, rather than a single task. Prominent large models include Codex~\cite{codex}, AlphaCode~\cite{AlphaCode}, SantaCoder~\cite{santacoder}, Starcoder~\cite{starcoder}, WizardCoder~\cite{wizardcoder}, InCoder~\cite{incoder}, CodeT5~\cite{codet5}, CodeGeeX~\cite{codegeex}, Code Llama~\cite{code_llama}, and Code-QWen~\cite{Qwen}. To improve the performance of code generation, researchers used optimized prompts~\cite{ChatGPT_Prompt, Prompt_Programming, Instruction_Tuning, Prompting_Is_Programming}, bring test cases~\cite{Generated_Tests} and collaborative roles~\cite{Selfcollaboration}. There are also some related studies on using large language models for other code tasks, such as dynamic programming~\cite{Dynamic}, compiler optimization~\cite{Compiler}, multilingual prompts~\cite{CodeFuse}, and program of thoughts~\cite{PoT} (PoT). 

\paragraph{Chain-of-Thought Prompting.}
To unleash the potential of LLMs~\cite{zhang2024mapneo,Liu2024E2LLMEA,que2024d,du2024chinese} in addressing complex reasoning tasks, chain-of-thought (CoT) prompting~\cite{cot,zero_cot} extends in-context learning with step-by-step reasoning processes, which handles complex reasoning tasks in the field of the code and mathematics by encouraging them to engage in step-by-step reasoning processes. Following this line of research, X-of-Thought (XoT) reasoning (CoT and its structural variants further)~\cite{xcot,tot,scot,got,owl,ji2024sevenllm,guo2024lemur} further expands the capabilities and applications of LLMs in complex reasoning and planning scenarios.
\paragraph{Intermediate Repersentation}
In the field of natural language processing, there exist many works using intermediate representation \cite{skeleton_aware_nmt,um4,m3p,low_resource_template,alm,soft_template,machine_created_language}, such as text generation and translation. The universal code is used as the intermediate representation, which typically omits details that are essential for the machine implementation of the algorithm. We perform the coarse-to-fine pattern for the code generation and translation, where the universal code first summarizes the algorithm process and then the programming language gives the accurate solution. The Unicode provides explicit help for code generation such as Chain-of-thought in LLM.

\section{Conclusion}
\label{sec:conclusion}
In this work, we put forth a state-of-the-art framework \ourmethod{} for both code translation and code generation. Using the universal code \uc{} as the intermediate representation, we effectively bridge different programming languages and facilitate code tasks.
In addition, we collect a dataset \instruct{} with 140K instruction instances from existing instruction datasets and the raw code snippets.
After being fine-tuned on \instruct{} with multi-task learning objectives, our model generates \uc{} and translates it into the final answer (executable code).
The evaluation results on code translation and generation tasks demonstrate that our method significantly improves the generalization ability, showing the efficacy and superiority of \ourmethod{}.

\section*{Limitations}
\label{sec:limitations}
We acknowledge the following limitations of this study: (1) The evaluation focuses on benchmark datasets (Humaneval, MBPP, and MultiPL-E), and the model's effectiveness in real-world programming scenarios or industry applications is not fully explored. (2) Our method is developed and evaluated primarily on programming language benchmarks. Its effectiveness in other domains or for non-programming-related tasks is not assessed, which limits the generalizability of our findings.

\section*{Acknowledege}
This work was supported in part by the National Natural Science Foundation of China (Grant Nos. U1636211, U2333205, 61672081, 62302025, 62276017), a fund project: State Grid Co., Ltd. Technology R\&D Project (ProjectName: Research on Key Technologies of Data Scenario-based Security Governance and Emergency Blocking in Power Monitoring System, Proiect No.: 5108-202303439A-3-2-ZN), the 2022 CCF-NSFOCUS Kun-Peng Scientific Research Fund and the Opening Project of Shanghai Trusted Industrial Control Platform and the State Key Laboratory of Complex \& Critical Software Environment (Grant No. SKLSDE-2021ZX-18).

\bibliography{custom}
\bibliographystyle{acl_natbib}

\end{document}